\documentclass[lettersize,journal]{IEEEtran}
\usepackage{amsmath,amsfonts}
\usepackage{algorithmic}
\usepackage{algorithm}
\usepackage{array}
\usepackage[caption=false,font=normalsize,labelfont=sf,textfont=sf]{subfig}
\usepackage{textcomp}
\usepackage{stfloats}
\usepackage{url}
\usepackage{verbatim}
\usepackage{graphicx}
\usepackage{cite}
\usepackage{booktabs}
\begin{document}

\title{Hyp-UML: Hyperbolic Image Retrieval with Uncertainty-aware Metric Learning}
\author{Shiyang Yan, Zongxuan Liu, Lin Xu
\thanks{Shiyang Yan is with Inria, France, Zongxuan Liu is with E.m. Data Technology Ltd., China, Lin Xu is with Xi'an Jiaotong University, China}}

\markboth{}%
{Shell \MakeLowercase{\textit{et al.}}: A Sample Article Using IEEEtran.cls for IEEE Journals}


\maketitle

\begin{abstract}
Metric learning plays a critical role in training image retrieval and classification. It is also a key algorithm in representation learning, e.g., for feature learning and its alignment in metric space. Hyperbolic embedding has been recently developed. Compared to the conventional Euclidean embedding in most of the previously developed models, Hyperbolic embedding can be more effective in representing the hierarchical data structure. Second, uncertainty estimation/measurement is a long-lasting challenge in artificial intelligence. Successful uncertainty estimation can improve a machine learning model's performance, robustness, and security. In Hyperbolic space, uncertainty measurement is at least with equivalent, if not more, critical importance. In this paper, we develop a Hyperbolic image embedding with uncertainty-aware metric learning for image retrieval. We call our method Hyp-UML: Hyperbolic Uncertainty-aware Metric Learning. Our contribution are threefold: we propose an image embedding algorithm based on Hyperbolic space, with their corresponding uncertainty value; we propose two types of uncertainty-aware metric learning, for the popular Contrastive learning and conventional margin-based metric learning, respectively. We perform extensive experimental validations to prove that the proposed algorithm can achieve state-of-the-art results among related methods. The comprehensive ablation study validates the effectiveness of each component of the proposed algorithm. 
\end{abstract}


\section{Introduction}
\IEEEPARstart{M}{etric} learning is a critical machine learning method which has massive applications, e.g., image retrieval~\cite{sohn2016improved, wang2019multi, musgrave2020metric,wang2019ranked, kim2020proxy}, clustering~\cite{xing2002distance}, transfer learning~\cite{ding2016robust}, few-shot learning~\cite{jiang2020multi}, person re-identification~\cite{hermans2017defense}, etc. The basic idea of metric learning is that the obtained distances between the data embedding must represent the semantic similarity of the corresponding data itself.

The choice of the embedding space in metric learning directly influences the metrics used for comparing representations. Typically,
embedding are transformed into a Hyper-sphere, i.e. the output
of the encoder is l2 normalized, resulting in using cosine or Euclidean
similarity as the distance measurement~\cite{hermans2017defense}. The community witnessed the great success of deep neural networks~\cite{he2016deep}, parameterized with millions of parameters also boosted via highly-optimized libraries~\cite{abadi2016tensorflow, paszke2017automatic}, theoretically having the capability to fit functions with arbitrary complexities. Nevertheless, even for this successful application of deep neural networks, data embeddings are in the Euclidean space, i.e., flat space with zero curvature~\cite{peng2021hyperbolic}. The embeddings in the Euclidean space were originally designed for grid data, which cannot provide meaningful geometrical representations for structural data, most often seen in real-world applications. A solution to represent the hierarchical geometrical of data is using the Hyperbolic space~\cite{peng2021hyperbolic, khrulkov2020hyperbolic}. This study proposes considering the Hyperbolic space for distance metric learning.

The unique property of Hyperbolic space is the
exponential volume growth concerning the radius of the data embedding in contrast to the polynomial growth in Euclidean space. This exponential property of the Hyperbolic space makes it especially suitable for embedding hierarchical data due to its increased representation power. The hierarchical data can be embedded in Poincar\' e's disk with arbitrarily low distortion. Most commonly seen data is intrinsically hierarchical, e.g., images with objects, natural language, knowledge graphs, etc. Hence, Hyperbolic embedding space is very suitable for most data. The other intriguing property of the Hyperbolic space is the capability to use relatively low-dimensional embedding without reducing the model's representation power.

Apart from the commonly seen structural property in real-world data, there is another critical problem in machine learning and artificial intelligence in general, i.e., uncertainty quantification. It is widely recognized that models developed using machine learning algorithms are widely applied for all types of inference and decision-making. It is vitally important to evaluate the reliability and efficacy
of the models before their application. Previous endeavours try to solve the uncertainty quantification problem from a perspective of Bayesian machine learning~\cite{wang2018adversarial, jospin2022hands,fernandez2022uncertainty}. Many approaches apply techniques like Markov chain Monte Carlo (MCMC)~\cite{gong2019icebreaker}, variational inference~\cite{wang2018adversarial}, Laplacian approximations~\cite{ritter2018scalable}. However, Bayesian approaches suggest a method to quantify uncertainty in neural networks by demonstrating all network parameters in a probabilistic structure~\cite{tomani2021towards}, which has a somewhat limited scope.

This paper quantifies the uncertainty in Hyperbolic metric learning embedding space. First, from the viewpoint of Hyperbolic geometry, all points of the Poincar\' e ball are equivalent. The methods we propose in this paper are considered hybrid as most of the model's intermediate layers use Euclidean operators, such as standard convolutions, fully-connected operations, or conventional attention mechanisms. At the same time, only the final embedding operates within the Hyperbolic geometry space. The hybrid nature of our model makes the origin (the centre of the Poincar\' e ball) a particular point. From the viewpoint of Euclidean embedding, the local volumes in the Poincar\' e ball expand exponentially from the origin to the boundary. The Poincar\' e ball forms a valuable property: the learned embeddings place more generic/ambiguous objects closer to the origin while moving more specific objects towards the boundary. Therefore, the distance to the origin in our models provides a natural estimate of uncertainty, i.e., the shorter the distance of an embedding to the origin, the more significant the uncertainty. This uncertainty estimation can be used in several ways, which will be discussed later.

This way of uncertainty estimation is justified for the following reasons. First, many existing vision architectures are designed to output data embeddings in the vicinity of zero (e.g., in the unit ball in the case of l2 normalized data). Another very critical property of Hyperbolic space (assuming the standard Poincar\' e ball model) is the existence of a reference point, i.e., the centre of the ball. We show that in image classification which constructs embeddings in the Poincar\' e model of Hyperbolic spaces the distance to the center can
serve as a measure of confidence in the model. Precisely, the input
images more familiar to the model get mapped
closer to the boundary. On the contrary, the images that confuse the model (e.g. images with significant noise or data samples from the unseen categories) are mapped closer to the centre.

With this Hyperbolic image embedding and their corresponding uncertainty measurement, we formulate a general adaptive metric learning method for image retrieval in this paper. First, we consider the Contrastive learning method for image retrieval~\cite{ermolov2022hyperbolic}. As~\cite{zhang2021temperature} suggested, the temperature is a critical parameter in the Contrastive learning loss, which can also reflect the uncertainty of the model's outputs. Hence, we follow the formation of uncertainty-aware Contrastive loss in~\cite{zhang2021temperature} and propose a novel uncertainty-aware Contrastive loss in the Hyperbolic space and its corresponding uncertainty estimation. Second, we consider the conventional metric learning loss, i.e., the representative Triplet loss~\cite{schroff2015facenet} in the Hyperbolic space. Given the inherent structural property of the Hyperbolic embedding, a fixed margin in the Triplet loss is not flexible enough for practical training. Hence, we introduce the estimated uncertainty value as the adaptive margin in the Triplet loss formulation, which shows a consistently improving effect. Our unified method is ``Hyp-UML'' (Hyperbolic Uncertainty-aware Metric Learning). In summary, our contributions are threefold:
\begin{itemize}
    \item We propose an image embedding algorithm for the widely used Transformer models based on Hyperbolic space, with their corresponding uncertainty measurement value.
    \item We propose two types of uncertainty-aware metric learning: popular Contrastive learning and conventional margin-based metric learning.
    \item We perform extensive experimental validations. The proposed algorithm achieves state-of-the-art results among related methods. The ablation study validates the effectiveness of each component of the proposed algorithm. 
\end{itemize}

\section{Related Work}
In this section, we review related methods and literature, including two subsections: Metric learning and Uncertainty representation, the two research fields most related to our proposed methods.
\subsection{Metric Learning}
\textbf{Deep Metric Learning} (DML) has been explored for several years in the deep learning field\cite{oh2016deep}. The main idea is to give a sampled dataset labelled for its identity as a training dataset to train a deep neural network. Given a set of untrained quarry samples during the evaluation and inference phases, the trained model is supposed to find the most similar samples in the gallery set for each quarry sample. The method can be applied to many computer vision tasks, including image retrieval, person re-identification\cite{yao2019deep}, vehicle re-identification, and face recognition\cite{deng2019arcface, schroff2015facenet}. The core of improving performance is to improve model representation ability that can make the model extract more accurate low-dimension embedding, ensuring samples from the same identity have lower distances and samples from different identities have larger distances. \cite{hadsell2006dimensionality} first proposed Contrastive loss that can map out-of-training samples to a low-dimension embedding and keep the invariant relationship with inputs. The main idea is to pull close the similar samples and push dissimilar ones away. Triplet loss\cite{schroff2015facenet} takes a triplet set that includes a positive, negative, and an anchor sample. The loss function aims to ensure that the distance between anchor and positive is less than that between anchor and negative in a given margin(generally 0.3).

\subsection{Hyperbolic Embedding}
\textbf{Hyperbolic space} \cite{peng2021hyperbolic, gromov1987hyperbolic} is a curve space in that the curvature is constantly negative. Because of its good hierarchical structure representation ability, the space has been noticed recently in the Graph Neural Network(GNN) field \cite{yang2022hyperbolic} and has been successfully applied in the recommendation system, knowledge graph. \cite{ganea2018hyperbolic}  first bridges the gap between Euclidean space DNN framework and Hyperbolic space. The paper uses the Poincar\' e model of Hyperbolic geometry to create a neural network representing Hyperbolic embedding. For NLP, \cite{gulcehre2018hyperbolic} first combines Hyperbolic space with attention to the mechanism by the Hyperboloid model, an unbounded Hyperbolic representation method for capturing complex hierarchical and power-law structures. For the Computer Vision tasks, \cite{khrulkov2020hyperbolic} proposed that Hyperbolic can also improve performance on image-related tasks like classification, representation learning, and few-shot learning. In addition, the paper suggests that the distance between feature points and origin can be treated as an uncertainty measurement that can be standard for how the input samples may be represented correctly. In our work, we take the Poincar\' e model as our Hyperbolic representation for its intuitive and simple optimization by gradient-based learning.

\subsection{Uncertainty Representation}
There are generally three types of uncertainty: approximation, aleatoric, and epistemic uncertainties~\cite{der2009aleatory}. First, approximation uncertainty describes the errors made by simplistic models unable to fit complex data. Second, aleatoric uncertainty (from the Greek word alea, meaning “rolling a dice”) accounts for the stochasticity of the data. Aleatoric uncertainty describes the variance of the conditional distribution of our target variable given our features. This uncertainty arises due to hidden variables or measurement errors and cannot be reduced by collecting more data under the same experimental conditions. Third, epistemic uncertainty (from the Greek word episteme, meaning “knowledge”) describes the errors associated with our model's lack of experience at some areas of the feature space. Therefore, epistemic uncertainty is inversely proportional to the density of training examples and could be reduced by collecting data in those low-density regions. As argued by Begoli et al.~\cite{begoli2019need}, uncertainty quantification is a problem of paramount importance when deploying machine learning models in sensitive domains such as information security~\cite{rakes2012security}, medical research~\cite{ranstam2009sampling}.

Despite its importance, uncertainty quantification is a largely unsolved problem. Prior literature on uncertainty estimation for deep neural networks is dominated by Bayesian methods~\cite{li2020uncertainty, molnar2022flow}, implemented in approximate ways to circumvent their computational intractability. Frequentist approaches on expensive ensemble models~\cite{egele2022autodeuq, abdar2021review}, which often cannot estimate asymmetric, multimodal predictive intervals.

\section{Methods}
In this section, we first introduce some necessary preliminary knowledge on hyperbolic space. Subsequently, we present our hyperbolic image embedding scheme for deep neural networks, i.e., the Transformer models in this paper. Last, we explain the proposed uncertainty measurement and the corresponding uncertainty-aware metric learning losses for the image retrieval task.

\begin{figure*}
    \centering
    \includegraphics[width=\linewidth]{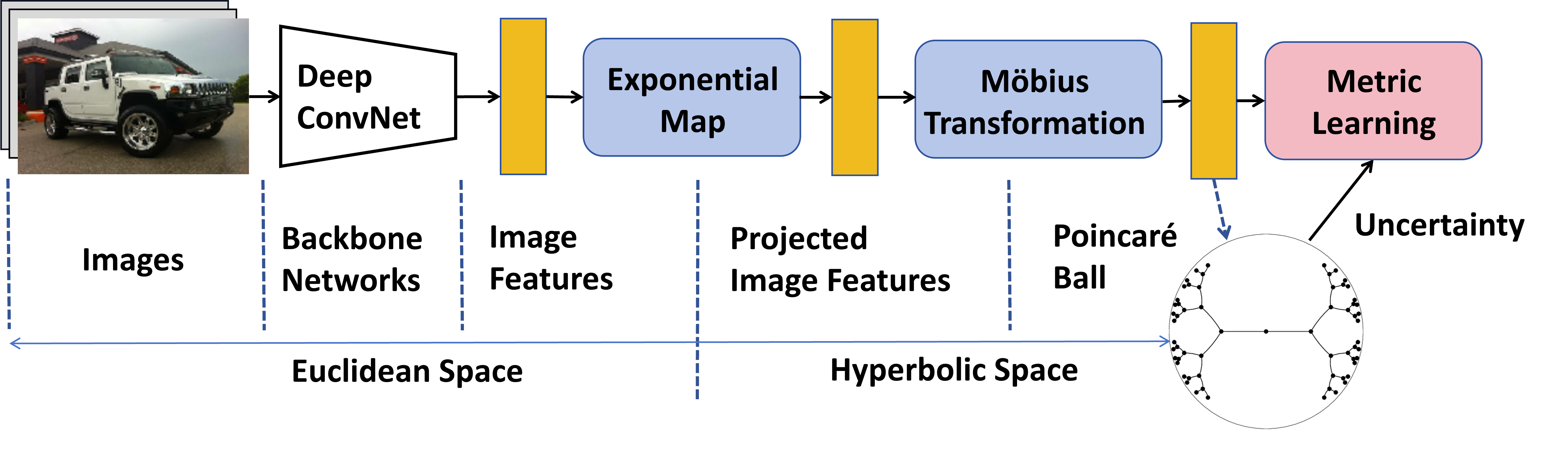}
    \caption{We test our method in classical image retrieval tasks. Given a set of images, we first apply different backbone deep ConvNets to extract image features, which are all in the Euclidean space. Subsequently, the image features are processed via exponential map and Möbius transformation to form a formal Poincar\' e ball Hyperbolic representation for the hierarchical structure. With the Hyperbolic embedding, we calculate the uncertainty value of the embedding and apply the Contrastive and Margin-based loss to perform end-to-end training of the neural networks for the image retrieval task. }
    \label{fig:system}
\end{figure*}

\begin{figure*}
    \centering
    \includegraphics[width=0.8\linewidth]{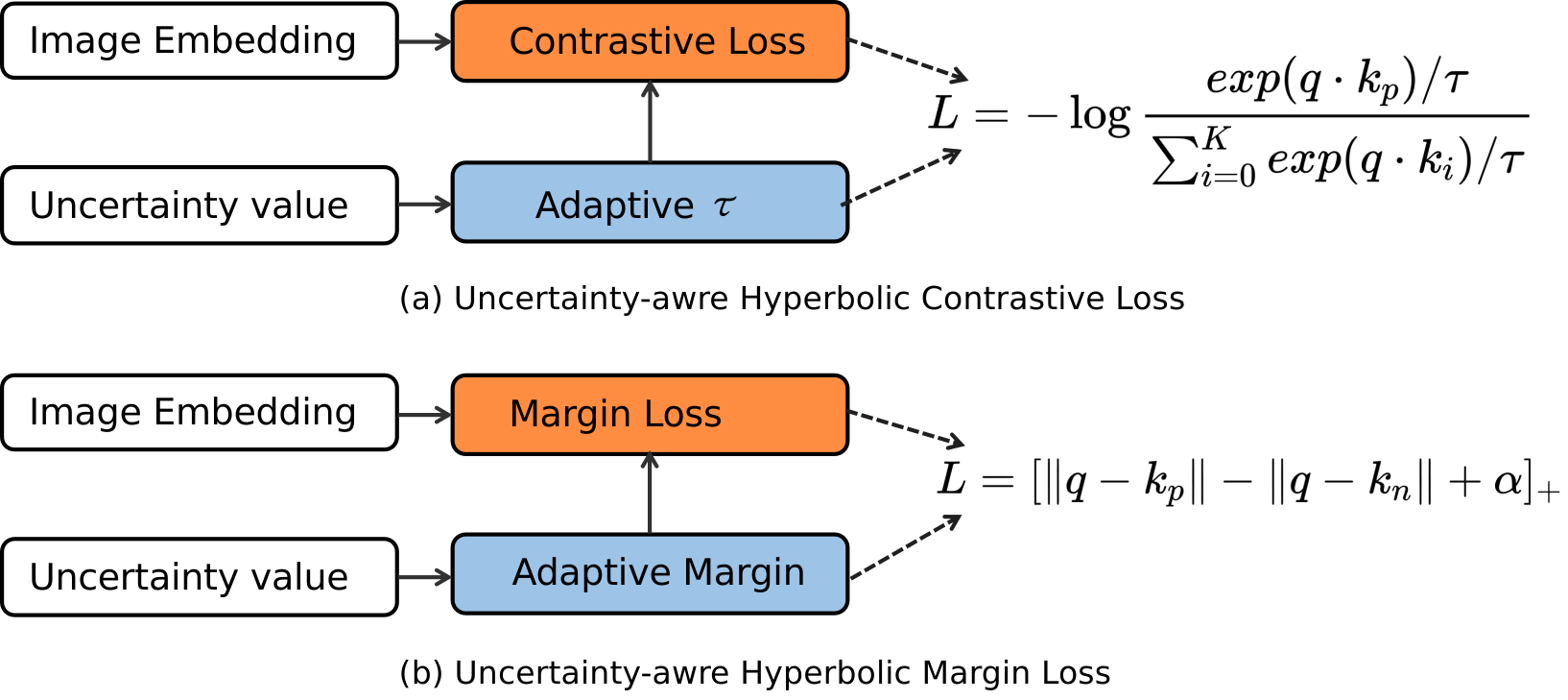}
    \caption{An illustration of the proposed two types of metric learning loss: (a) Adaptive $\tau$ Contrastive loss and (b) Adaptive Margin metric learning loss. In the equations, $q$ is the query sample, $k_p$ is the positives, and $k_n$ is the negatives. $K$ is the size of one training batch.}
    \label{fig:losses}
\end{figure*}

\subsection{Preliminaries on Hyperbolic space}
The n-dimensional Hyperbolic space $\mathbf{H}^n$ is a Riemannian manifold of constant negative curvature. The property of constant negative curvatures makes it similar to the ordinary Euclidean sphere, which has constant positive curvature. The Hyperbolic space is very different from that of Euclidean. It is known that Hyperbolic space cannot be isometrically embedded into Euclidean space. There are several well-studied models of Hyperbolic geometry. Among them, the Poincar\' e ball is widely known.

\paragraph{Poincar\' e ball model.} The Poincar\' e ball model ($\mathbf{D}^n, g^{\mathbf{D}}$) is defined by the manifold,
\begin{equation}
    \mathbf{D}^n = \{ x \in  \mathbf{R}^n: \Vert x \Vert < 1 \},
\end{equation}
endowed with the Riemannian metric $g^{\mathbf{D}}(x) = \lambda_x^2 g^E$, where $\lambda_x = \frac{1}{1 - \Vert x \Vert^2}$ is a conformal factor and $g^E$ is the euclidean metric tensor $g^E = \mathbf{I}^n$.

\paragraph{Möbius Addition.}
Hyperbolic spaces are not Euclidean vector spaces. In order to perform operations such as addition, we need to introduce a so-called gyrovector formalism. For a pair $x, y \in \mathbf{D}_c^n$, their addition is defined as:
\begin{equation}
x\oplus_cy = \frac{(1+2c\langle x,y\rangle + c\Vert y \Vert)^2)x + (1 - c\Vert x \Vert^2)y)}{1 + 2c\langle x, y \rangle + c^2\Vert x \Vert^2 \Vert y \Vert^2}.
\end{equation}

\paragraph{Hyberbolic Distance.}
The Hyperbolic distance between $x, y\in \mathbf{D}_c^n$ is defined in the following expression:
\begin{equation}
    D_{Hyp}(x, y) = \frac{1}{\sqrt{c}}arctanh(\sqrt{c} \Vert - x \oplus_c y \Vert),
\end{equation}
where when $c\rightarrow 0$ the distance reduces to euclidean distance, i.e., $\lim_{c\rightarrow0} D_{Hyp}(x, y) = 2\Vert x-y \Vert$.

\paragraph{Exponential Map.}
There is a bijection mapping from Euclidean space to the Poincar\' e model of Hyperbolic geometry. This mapping is termed exponential, while its inverse mapping is called logarithmic. The mapping can be defined as for some fixed base point $x\in \mathbf{D}_c^n$; the exponential mapping is a function $exp_x^c: \mathbf{R^n} \rightarrow \mathbf{D}_c^n$, defined as:
\begin{equation}
    exp_x^c(v) = x\oplus_c(tanh(\sqrt{c}\frac{\lambda_x^c\Vert v \Vert}{2}) \frac{v}{\sqrt{c}\Vert v \Vert}),
\end{equation}
where the base point x is usually set to 0, which makes the formulas and implementation easier.

\subsection{Uncertainty measurement for Hyperbolic embedding}
As discussed in~\cite{khrulkov2020hyperbolic}, in image classification, which constructs embeddings in the Poincar\' e model of Hyperbolic spaces, the distance to the centre can
serve as a measure of confidence in the model, i.e., the input
images which are more familiar to the model get mapped
closer to the boundary, and images that confuse the model
(e.g., blurry or noisy images, instances of an unfamiliar hard image) are mapped closer to the centre. In a word, the certainty value is the distance of the embedding to the centre. Given a data point $x$, the uncertainty can be expressed as,
\begin{equation}
\begin{split}
    & D_{Hyp}(x, 0) = \frac{1}{\sqrt{c}}arctanh(\sqrt{c} \Vert - x \oplus_c 0 \Vert), \\
    & Uncertainty(x) = 1 - D_{Hyp}(x, 0),
\end{split}
\end{equation}

\subsection{Uncertainty-aware metric learning}
This section introduces the proposed uncertainty-aware metric learning losses, including Hyperbolic Contrastive loss and conventional margin-based loss. A schematic illustration of the proposed scheme is presented in Figure~\ref{fig:losses}. In short, the image embedding and the corresponding uncertainty value are formulated via our Hyperbolic embedding modules and finally trained via the proposed losses. 
\paragraph{Hyperbolic Contrastive Loss}
We give a brief overview of commonly applied metric learning loss, i.e. Contrastive learning loss, which is expressed as, 

\begin{equation}
\begin{split}
   & \mathcal{L}_{Contrastive} \\
    = & \mathbf{E}_{t, t', t_{1:B}\sim p(t)}
    \mathbf{E}_{x_{1:B}\sim p(t)}
\left [ \log \frac{e^{f(t(x_i))\dot f(t'(x_i))/\tau}}{\frac{1}{k}\sum_{j\in{1:B}}e^{f(t(x_i))\dot f(t_j(x_j))/\tau}} \right ], \\
& \tau = \log \left (\frac{Uncertainty(x_i)}{Uncertainty(x_{i\sim B}).max()
} + 1 \right )/scale, 
\end{split}
\end{equation}
where $x_{1:B} = \{x_1, ..., x_B\}$ represents $B$ i.i.d. samples in a batch, and $\tau$ is the temperature. $f(*)$ is the embedding function. We call the transformation (data augmentation)($t(*)$) of the same image ($x_i,x_i$) ``positive examples" and the transformation of different images ($x_i,x_j$) ``negative examples". $scale$ is a hyper-parameter that determines the numerical scale of the uncertainty estimated value.

The role of temperature in Contrastive learning loss is critical, as reported in many previous approaches. $\tau$ is to scale the sensitivity of the loss function. A $\tau$ closer to 0 would accentuate when representations are different, resulting in a more significant gradient, and vice versa~\cite{9577669,zhangTemp2021}. The sensitivity of $tau$ naturally represents a measure of representation uncertainty.

Unlike previous methods, which explicitly learn a function representing $\tau$, we apply Hyperbolic embedding with uncertainty measurement combined with Contrastive learning loss. In other words, the value of $tau$ is set as the distance of the embedding to the centre of the Poincar\' e ball.

\paragraph{Hyperbolic Margin-based Loss}
Recall margin-based metric learning losses; among them, Triplet loss~\cite{hermans2017defense} is a classical and stable margin-based loss. 

The parameter margin plays a key role in margin-based loss functions. A large margin split the positive and negative examples much further but with less stable training. This paper applies the uncertainty value as an explicit dynamic margin parameter in Triplet loss. The idea is that with more uncertain samples, we should split much ``harder'' to separate the positive and negative samples and vice versa. By doing this, the Hyperbolic embedding is seamlessly combined with margin-based loss. The dynamic margin is empowered by uncertainty measurement. Thus the training is more effective, which is expressed as,
\begin{equation}
\begin{split}
& \mathcal{L}_{Triplet} = \sum_{i=1}^B [a_is(e^{a}_i, e^{n}_i) - a_is(e^{a}_i, e^{p}_i) + \alpha]_+, \\
& \alpha = \log \left( \frac{Uncertainty(e^{a}_i)}{Uncertainty(e^{a}_{i\sim B}).max()} +1 \right ) * \exp (Curvature)),
\end{split}
\end{equation}
where  $e^a_i$ is the anchor examples in Triplet loss, $e^p_i$ is the positive samples and the $e^n_i$ is the negative samples. $\alpha$ is the margin parameter, which is set as the same value of uncertainty measurement of the anchor example.

\paragraph{System Architecture}

The whole system architecture is shown in Figure~\ref{fig:system}. We deploy our method in classical image retrieval tasks. Given a set of images, we first apply different backbone deep ConvNets to extract image features, which are all in the Euclidean space. Subsequently, the image features are processed via the exponential map, Möbius transformation, to form a formal Poincar\' e ball Hyperbolic representation for the hierarchical structure. With the Hyperbolic embedding, we calculate the uncertainty value of the embedding and apply the Contrastive and Margin-based loss to perform end-to-end training of the neural networks for the image retrieval task. 

Note that from the inputs to the image features from backbone networks, the embedding space is all in the Euclidean space, the same as the conventional deep neural networks training pipeline. The image embedding is further processed via the Hyperbolic embedding with several additional layers. Hence, the computational complexity is controlled in limited scope, i.e., the computing time is not increased significantly, but with higher retrieval performance due to the hierarchical representing capacity of the Hyperbolic embedding.

\section{Experiments}
This section reports our comprehensive experimental validation for the proposed methods. The section includes the basic setup of the experiments, datasets introduction, backbone models, training and testing implementation details, the numerical evaluation results, and qualitative evaluations. 
\subsection{Experiments Setup}
This section will demonstrate our experiment implementation and the running detail settings. To fully evaluate our proposed methods, we conducted our experiment on several datasets, including CUB-200-2011, Cars196, Standford Online Product (SOP), and In-shop Clothes Retrieval (In-shop). We take the Transformer model as our main backbone for extracting representation embeddings, which can show that the proposed method has comparable generalization capability.
\subsubsection{Datasets}
\paragraph{CUB-200-2011~\cite{krause20133d}} is a dataset that includes 11,788 images of 200 bird species. The dataset can be used for classification, object detection, and image retrieval. In this work, we take the dataset as the training dataset for image retrieval. Following the general setting, we split the first 100 classes, including 5864 images as a training set and the remaining classes and images for testing. \paragraph{Cars196~\cite{krause20133d}} is a widely used dataset for vehicle retrieval. It contains 16185 images of 196 classes of cars. We take the first 98 classes, 8,054 images for training, and the remaining 98 classes, 8131 images for testing.  \paragraph{Stanford Online Product~\cite{oh2016deep}} is a dataset published by Stanford University that obtained online product images from eBay.com. The dataset has 120,053 images of 22634 classes. As the standard setting, we divide 59551 images of 11,318 classes into the training set and 60,502 images of 11,316 classes into the testing set. \paragraph{In-shop Clothes Retrieval~\cite{liu2016deepfashion}} is a subset of the DeepFashion dataset. It is specialized as an image retrieval benchmark comprising 7,986 clothing items for training and testing. As mentioned in \cite{ermolov2022hyperbolic}, we take 3,997 items for training and the other 3985 for testing.

\subsubsection{Backbone Models}
\textbf{ResNet50}\cite{he2016deep} is a full convolution network. So, we take it as a typical CNN-based model for getting embeddings. We remove the last classification layer and get the embedding from the backbone directly, whose size is 2048-d. We all take their basic model configuration for all transformer-based models, including \textbf{VIT}, \textbf{DEIT}, and \textbf{BEIT}, which means they share a similar architecture. The patch size is 16, the head number is 12, and the depth is 12. We only fix the first patch embedding layer during training; the other layers attend the training process. We remove the original classification head and get a 768-d embedding from the backbone. The CNN-based and transformer-based models are pre-trained on the ImageNet classification dataset.

The embedding from the backbones is then sent to a Hyperbolic representation head. The head consists of a linear layer that converts embedding to 128-d and a Hyperbolic projection layer that projects the 128-d Euclidean space embedding to 128-d Hyperbolic space embedding. To improve the training efficiency, the projection layer also takes a gradient clip strategy proposed by\cite{guo2022clipped}.

\begin{table*}[ht]
\centering
\caption{The configuration of training details on different datasets.}\label{Tab:configuration}
\begin{tabular}{l|ll|ll|ll|ll}
\toprule
Datasets       & \multicolumn{2}{|c}{CUB200} & \multicolumn{2}{|c}{CARS196}& \multicolumn{2}{|c}{SOP} & \multicolumn{2}{|c}{In-shop}  \\
\hline
Epochs      &  50 & 150 & \multicolumn{2}{c}{300} & \multicolumn{2}{|c}{200} & \multicolumn{2}{|c}{400}  \\
\hline
Methods & Contrastive & Triplet & Contrastive & Triplet & Contrastive & Triplet & Contrastive & Triplet \\
\hline
Batch size & 300 & 200 & 294 & 194 & 294 & 200 & 294 & 200 \\
\hline 
Sample num & 3 & 2 & 3 & 2 & 3 & 2 & 3 & 2  \\
\bottomrule
\end{tabular}
\end{table*}
\subsubsection{Training and Testing Settings}
We conducted experiments on Contrastive and triplet loss, respectively, to ensure the proposed method is applicable on different criteria. For Contrastive loss, we set the learning rate $3e{-5}$ for all methods. We test uncertainty on curvature 0.05, 0.1, and 0.3. We use a different scale for different curvature values to get the best performance. Depending on different datasets, we use different training epochs; more details are shown in Table 1. 

We use Adam as an optimizer, and the weight decay is 0.01. All input images are resized to 256. Automatic Mixed Precision optimizes all training processes in O2 mode. For evaluation, we can calculate evaluation metrics in the cumulated matching characteristics (CMC) curve. The CMC curve is a widely used evaluation metric that shows the probability that a query identity appears in different-sized candidate lists. CMC top-k accuracy would be if top-k ranked gallery samples contain the query identity. Otherwise, the accuracy is 0. 

\begin{table*}[ht]
\centering
\caption{Comparison of the proposed methods with other State-of-the-arts.}\label{Tab:sota}
\resizebox{\linewidth}{!}{ 
\begin{tabular}{l|llll|llll|llll|llll}
\toprule
Datasets       & \multicolumn{4}{c}{CUB200} & \multicolumn{4}{c}{CARS196}& \multicolumn{4}{c}{SOP} & \multicolumn{4}{c}{In-shop}  \\
\hline
Method     & 1          & 2          & 4          & 8          & 1           & 2         &  4        & 8          & 1 & 10 & 100 & 1000 & 1 & 10 & 20 & 30 \\
\hline
Margin~\cite{8237571} & 63.9 & 75.3 & 84.4&  90.6 & 79.6 & 86.5 & 91.9 & 95.1 & 72.7 & 86.2 & 93.8 & 98.0 & - & - & - & - \\
FastAP~\cite{cakir2019deep} & - & - & - & - & - & - & - & - & 73.8 & 88.0 & 94.9 & 98.3 & - & - & - & - \\
NSoftmax~\cite{zhai2018classification} & 56.5 & 69.6 & 79.9 & 87.6 & 81.6 & 88.7 & 93.4 & 96.3 & 75.2 & 88.7 & 95.2 & - & 86.6 & 96.8 & 97.8 & 98.3 \\
MIC~\cite{roth2019mic} & 66.1 & 76.8 & 85.6 & - & 82.6 & 89.1 & 93.2 & 77.2 & 89.4 & 94.6 - & 88.2 & 97.0 & - & 98.0 \\
XBM~\cite{wang2020cross} & - & - & - & - & - & - & - & - & 80.6 & 91.6 & 96.2 & 98.7 & 91.3 & 97.8 & 98.4 & 98.7 \\
IRT\_{R}~\cite{el2021training} & 72.6 & 81.9 & 88.7 & 92.8 & - & -  & - & - & 83.4 & 93.0 & 97.0 & 99.0 & 91.1 & 98.1 & 98.6 & 99.0 \\
Sph-DeiT~\cite{ermolov2022hyperbolic} & 73.3 & 82.4 & 88.7 & 93.0 & 77.3 & 85.4 & 91.1 & 94.4 & 82.5 & 93.1 & 97.3 & 99.2 & 89.3 & 97.0 & 97.9 & 98.4 \\
Sph-DINO~\cite{ermolov2022hyperbolic} & 76.0 & 84.7 & 90.3 & 94.1 & 81.9 & 88.7 & 92.8 & 95.8 & 82.0 & 92.3 & 96.9 & 99.1 & 90.4 & 97.3 & 98.1 & 98.5 \\
Sph-ViT (ImageNet-21k)~\cite{ermolov2022hyperbolic} & 83.2 & 89.7 & 93.6 & 95.8 & 78.5 & 86.0 & 90.9 & 94.3 & 82.5 & 92.9 & 97.4 & 99.3 & 90.8 & 97.8 & 98.5 & 98.8 \\
Hyp-DeiT~\cite{ermolov2022hyperbolic} & 74.7 & 84.5 & 90.1 & 94.1 & 82.1 & 89.1 & 93.4 & 96.3 & 83.0 & 93.4 & 97.5 & 99.2 & 90.9 & 97.9 & 98.6 & 98.9 \\
Hyp-DIDO~\cite{ermolov2022hyperbolic} & 78.3 & 86.0 & 91.2 & 94.7  & 86.0 & 91.9 & 95.2  & 97.2 & 84.6 & 94.1 & 97.7 & 99.3 & 92.6 & 98.4 & 99.0 & 99.2 \\
Hyp-ViT (ImageNet-21k)~\cite{ermolov2022hyperbolic} & 84.0 & 90.2 & 94.2 & 96.4 & 82.7 & 89.7 & 93.9 & 96.2 & 85.5 & 94.9 & 98.1 & 99.4 & 92.7 & 98.4 & 98.9 & 99.1 \\
\hline
Hyp-ViT adaptive margin & 0.857 & 0.91	 & 0.942 & 0.963 & 0.742 & 0.841 & 0.908 & 0.951 & 0.861 & 0.952 & 0.983 & 0.996 & 0.933 & 0.988 & 0.992 & 0.994\\
Hyp-DeiT adaptive margin & 0.702 & 0.81	& 0.885	& 0.933 & 0.623 & 0.738 & 0.835 & 0.907 &  0.804 & 0.922 & 0.971 & 0.992 & 0.899 & 0.979 & 0.986 & 0.99\\
Hyp-BeiT adaptive margin & 0.875 & 0.922 & 0.948 & 0.964 & 0.747 & 0.844 & 0.915 & 0.954 & 0.869 & 0.957 & 0.985 & 0.996 & 0.943 & 0.99 & 0.993 & 0.995 \\
\hline
Ours: Hyp-ViT adaptive $\tau$ & 0.853 & 0.913 & 0.941 & 0.961 & 0.865 & 0.923 & 0.954 & 0.974 & 0.833 & 0.937 & 0.976 & 0.993 & 0.924 & 0.985 & 0.99 & 0.992\\
Ours: Hyp-DeiT adaptive $\tau$ & 0.758 & 0.851 & 0.908 & 0.945 & 0.856 & 0.92 & 0.956 & 0.977 & 0.821 & 0.931 & 0.974 & 0.992 & 0.92 & 0.983 & 0.988	& 0.991\\
Ours: Hyp-BeiT adaptive $\tau$ & \pmb{0.883} & \pmb{0.923} & \pmb{0.947} & \pmb{0.963} & \pmb{0.916} & \pmb{0.954} & \pmb{0.971} & \pmb{0.984} & \pmb{0.857} & \pmb{0.95} & \pmb{0.982} & \pmb{0.995} & \pmb{0.943} & \pmb{0.988} & \pmb{0.992} & \pmb{0.994} \\
\bottomrule
\end{tabular}}
\end{table*}

\begin{table*}[ht]
\caption{The table compares the Hyperbolic embedding metric learning results between traditional Contrastive loss without uncertainty and our combined uncertainty methods under different curvature}\label{Tab:Contrastive}
\centering
\resizebox{\linewidth}{!}{  
\begin{tabular}{l|llll|llll|llll|llll}
\toprule
Datasets       & \multicolumn{4}{c}{CUB200} & \multicolumn{4}{c}{CARS196}& \multicolumn{4}{c}{SOP} & \multicolumn{4}{c}{In-shop}  \\
\hline
Methods (Contrastive Loss)      & 1 & 2 &  4 &  8 & 1 & 2 & 4 & 8 & 1 & 10 & 100 & 1000 & 1 & 10 & 20 & 30 \\
\hline
Curvature 0.05  & \multicolumn{4}{c}{scale 0.22}&  \multicolumn{4}{|c}{scale 0.02}&  \multicolumn{4}{|c}{scale 0.02} &  \multicolumn{4}{|c}{scale 0.02}  \\
\hline
Hyp ViT base fix $\tau$ 0.2 & 0.85	& 0.911	& 0.942	& 0.961 & 0.853 & 0.915 & 0.949 & 0.971 & 0.829 &	0.933 &	0.975 &	0.993 & 0.92 & 0.984 & 0.989 & 0.991\\
Hyp ViT base uncertainty as $\tau$ & 0.853 & 0.911 & 0.94	& 0.96 & 0.863 &	0.924 &	0.956 &	0.975 & 0.833 &	0.937 &	0.976 &	0.992 & 0.926 & 0.984 & 0.989 & 0.991\\

Hyp DeiT base fix $\tau$ 0.2 & 0.752	& 0.848	& 0.905	& 0.946 & 0.84 & 0.907 & 0.944 & 0.966 & 0.818 & 0.928 & 0.973 & 0.992 & 0.922 & 0.982 & 0.988 &	0.991\\
Hyp DeiT base uncertainty as $\tau$ & 0.758 & 0.852 & 0.907 &0.95 & 0.855 & 0.918 & 0.955 & 0.976 & 0.82 & 0.93 & 0.973 & 0.992 & 0.918 & 0.983 & 0.989	& 0.991\\

Hyp BeiT base uncertainty as $\tau$ & \pmb{0.883} & \pmb{0.923} & \pmb{0.947} & \pmb{0.963} & \pmb{0.916} & \pmb{0.954} & \pmb{0.971} & \pmb{0.984} &  \pmb{0.857} &  \pmb{0.95} &  \pmb{0.982} &  \pmb{0.995} &  \pmb{0.943} &  \pmb{0.988} &  \pmb{0.992} &  \pmb{0.994} \\
\hline
Curvature 0.1  & \multicolumn{4}{c}{scale 0.22}&  \multicolumn{4}{|c}{scale 0.02}&  \multicolumn{4}{|c}{scale 0.02} &  \multicolumn{4}{|c}{scale 0.02}  \\
\hline
Hyp ViT base fix $\tau$ 0.2 & 0.851 & 0.911 & 0.942 & 0.962 & 0.855 & 0.918 & 0.951 & 0.97 & 0.825 & 0.931 & 0.974 & 0.992 & 0.921 & 0.982 & 0.988 & 0.991\\
Hyp ViT base uncertainty as $\tau$ & 0.853 & 0.913 & 0.941 & 0.961 & 0.865 & 0.923 & 0.954 & 0.974 & 0.833 & 0.937 & 0.976 & 0.993 & 0.924 & 0.985 & 0.99 & 0.992\\

Hyp DeiT base fix $\tau$ 0.2 & 0.75 & 0.846 & 0.906 & 0.945 & 0.846 & 0.906 & 0.945 & 0.968 & 0.816 & 0.926 & 0.972 & 0.921 & 0.982 & 0.989 & 0.991\\
Hyp DeiT base uncertainty as $\tau$ & 0.758 & 0.851 & 0.908 & 0.945 & 0.856 & 0.92 & 0.956 & 0.977 & 0.821 & 0.931 & 0.974 & 0.992 & 0.92 & 0.983 & 0.988	& 0.991\\

Hyp BeiT base uncertainty as $\tau$ & \pmb{0.88} & \pmb{0.924}	& \pmb{0.946}	& \pmb{0.962} & \pmb{0.912} & \pmb{0.951} & \pmb{0.97} & \pmb{0.983} &  \pmb{0.854}	&  \pmb{0.95} &  \pmb{0.981} &  \pmb{0.994} &  \pmb{0.942} &  \pmb{0.987} &  \pmb{0.992} &  \pmb{0.994} \\
\hline
Curvature 0.3  & \multicolumn{4}{c}{scale 0.22}&  \multicolumn{4}{|c}{scale 0.12}&  \multicolumn{4}{|c}{scale 0.12} &  \multicolumn{4}{|c}{scale 0.12}  \\
\hline
Hyp ViT base fix $\tau$ 0.2 &0.849	& 0.912	& 0.943	& 0.962& 0.856 & 0.919 & 0.952 & 0.973 & 0.809 & 0.921 & 0.971 & 0.991 & 0.914 & 0.979 & 0.985 & 0.988\\
Hyp ViT base uncertainty as $\tau$ & 0.846 & 0.906 & 0.938 & 0.959 & 0.871 & 0.926 & 0.957 & 0.976 & 0.816 & 0.928 & 0.972 & 0.992 & 0.921 & 0.982 & 0.987 & 0.99\\

Hyp DeiT base fix $\tau$ 0.2 &  0.755 & 0.847 & 0.907 & 0.944 & 0.842 & 0.907 & 0.946 & 0.97 & 0.797 & 0.917 & 0.968 & 0.991 & 0.913 & 0.979 & 0.986 & 0.989 \\
Hyp DeiT base uncertainty as $\tau$ & 0.756 & 0.849 & 0.905 & 0.945 & 0.861 & 0.919 & 0.952 & 0.975 & 0.808 & 0.922 & 0.971 & 0.992 & 0.913 & 0.98 & 0.988	& 0.99\\

Hyp BeiT base uncertainty as $\tau$ & \pmb{0.879} & \pmb{0.923} & \pmb{0.947} & \pmb{0.963} &  \pmb{0.905} &  \pmb{0.926} &  \pmb{0.957} &  \pmb{0.976} &  \pmb{0.846} &  \pmb{0.944} &  \pmb{0.98} &  \pmb{0.994} &  \pmb{0.936} &  \pmb{0.987} &  \pmb{0.991}	&  \pmb{0.993} \\
\bottomrule
\end{tabular}}
\end{table*}

\begin{table*}[ht]
\caption{The table compares different $\tau$ value on the CUB200 datasets}\label{Tab:abs_tau}
\centering 
\begin{tabular}{l|llll|}
\toprule
Datasets       & \multicolumn{4}{c}{CUB200}   \\
\hline
Methods (Contrastive Loss)      & 1 & 2 &  4 &  8 \\
\hline
ViT $\tau$=0.4 & 0.85 & 0.907 & 0.94 & 0.961 \\
ViT $\tau$=0.6 & 0.827 & 0.888 & 0.925 & 0.952 \\
ViT $\tau$=0.8 & 0.807 & 0.872 & 0.912 & 0.946 \\
ViT $\tau$=0.2 & 0.849 & 0.913 & \textbf{0.943} & \textbf{0.963} \\
\hline
ViT uncertainty as $\tau$ & \textbf{0.853} & \textbf{0.913} & 0.941 & 0.961  \\
\bottomrule
\end{tabular}
\end{table*}

\begin{table*}[ht]
\caption{The table compares the Hyperbolic embedding  metric learning results between traditional triplet loss with fixed margin 0.3 and our uncertainty as dynamic margin methods under curvature 0.05}\label{Tab:triplet}
\centering
\resizebox{\linewidth}{!}{
\begin{tabular}{l|llll|llll|llll|llll}
\toprule
Datasets       & \multicolumn{4}{c}{CUB200} & \multicolumn{4}{c}{CARS196}& \multicolumn{4}{c}{SOP} & \multicolumn{4}{c}{In-shop}  \\
\hline
Methods (Triplet Loss)      & 1          & 2          & 4          & 8          & 1           & 2         &  4        & 8          & 1 & 10 & 100 & 1000 & 1 & 10 & 20 & 30 \\
\hline
\multicolumn{17}{c}{Curvature 0.05} \\
\hline
Hyp ViT base fix margin 0.3  & 0.852 & 0.911 & 0.943 & 0.963 & 0.603 & 0.729 & 0.828 & 0.902 & 0.858 & 0.951 & 0.983 & 0.996 & 0.931 & 0.987 & 0.992 & 0.994\\
Hyp ViT base uncertainty as margin & 0.859 & 0.914 & 0.944 & 0.963 & 0.623 & 0.745 & 0.839 & 0.908 & 0.861	& 0.952	& 0.984	& 0.996 & 0.934 & 0.988 & 0.992 & 0.994\\
Hyp DeiT base fix margin 0.3  & 0.696 & 0.808 & 0.885 & 0.932 & 0.6 & 0.717 & 0.82 & 0.9 & 0.796 & 0.917 & 0.969 & 0.991 & 0.889 & 0.976 & 0.984 & 0.989\\
Hyp DeiT base uncertainty as margin & 0.703 & 0.811 & 0.886 & 0.933 & 0.603 & 0.719 & 0.823 & 0.899 &  0.804	& 0.921	& 0.971	& 0.992  & 0.9	& 0.979	& 0.986	& 0.99\\
Hyp BeiT base uncertainty as margin & \pmb{0.875} & \pmb{0.922} & \pmb{0.948} & \pmb{0.964} & \pmb{0.747} & \pmb{0.844} & \pmb{0.915} & \pmb{0.954} & \pmb{0.869} & \pmb{0.957} & \pmb{0.985} & \pmb{0.996} & \pmb{0.943} & \pmb{0.99} & \pmb{0.993} & \pmb{0.995} \\
\hline
\multicolumn{17}{c}{Curvature 0.1} \\
\hline
Hyp ViT base fix margin 0.3  & 0.85	& 0.909	& 0.941	& 0.962 & 0.654 & 0.775 & 0.866 & 0.927 & 0.857 & 0.951 & 0.984 & 0.995 & 0.929 & 0.987 & 0.992 & 0.994\\
Hyp ViT base uncertainty as margin & 0.86	& 0.914	& 0.944	& 0.962 & 0.665 & 0.778 & 0.867 & 0.93 & 0.861 & 0.952	& 0.984	& 0.996 & 0.933	& 0.987	& 0.993	& 0.994\\
Hyp DeiT base fix margin 0.3  & 0.697 & 0.805 & 0.883 & 0.931 & 0.617	& 0.735	& 0.831	& 0.905 & 0.796	& 0.917	& 0.969	& 0.991 & 0.885	& 0.977	& 0.984	& 0.988\\
Hyp DeiT base uncertainty as margin & 0.702 & 0.81	& 0.885	& 0.933 & 0.623 & 0.738 & 0.835 & 0.907 &  0.804 & 0.922 & 0.971 & 0.992 & 0.899 & 0.979 & 0.986 & 0.99\\
Hyp BeiT base uncertainty as margin & \pmb{0.874} & \pmb{0.921} & \pmb{0.949} & \pmb{0.964} & \pmb{0.742} & \pmb{0.839} & \pmb{0.905} & \pmb{0.952} & \pmb{0.87}	& \pmb{0.957}	& \pmb{0.986}	& \pmb{0.996} & \pmb{0.942}	& \pmb{0.99} & \pmb{0.993} & \pmb{0.995}\\
\hline
\multicolumn{17}{c}{Curvature 0.3} \\
\hline
Hyp ViT base fix margin 0.3  & 0.849 & 0.91	& 0.942	& 0.962 & 0.729 & 0.829 & 0.903 & 0.948 & 0.86 & 0.952 & 0.983 & 0.995 & 0.933 & 0.988 & 0.992 & 0.994\\
Hyp ViT base uncertainty as margin & 0.857 & 0.91	 & 0.942 & 0.963 & 0.742 & 0.841 & 0.908 & 0.951 & 0.861 & 0.952 & 0.983 & 0.996 & 0.933 & 0.988 & 0.992 & 0.994\\
Hyp DeiT base fix margin 0.3  & 0.696 & 0.803 & 0.881 & 0.93& 0.56 & 0.681 & 0.783 & 0.87& 0.798 & 0.918 & 0.969 & 0.991 & 0.887 & 0.978 & 0.985 & 0.988\\
Hyp DeiT base uncertainty as margin & 0.699 & 0.807 & 0.883 & 0.933 & 0.56 & 0.685 & 0.787 & 0.872 & 0.805 & 0.922 & 0.971 & 0.992 & 0.9	& 0.979	& 0.986	& 0.989\\
Hyp BeiT base uncertainty as margin & \pmb{0.868} & \pmb{0.918} & \pmb{0.947} & \pmb{0.964} & \pmb{0.682} & \pmb{0.791} & \pmb{0.873} & \pmb{0.931} & \pmb{0.869} & \pmb{0.957} & \pmb{0.985} & \pmb{0.996} & \pmb{0.942}	& \pmb{0.99} & \pmb{0.993} & \pmb{0.995}\\
\bottomrule
\end{tabular}} 
\end{table*}

\subsection{Experimental results}

\begin{table*}[ht]
\centering
\caption{The effect of uncertainty awareness of Contrastive learning for different models on CARS196 dataset}\label{Tab:testcase}
\begin{tabular}{l|llll}
\toprule
Datasets       & \multicolumn{4}{c}{CARS196}   \\
\hline
& 1          & 2          & 4          & 8          \\
\hline
Hyp ViT small fix $\tau$       & 0.799 & 0.877 & 0.925 & 0.957 \\
Hyp ViT small uncertainty as $\tau$  & 0.826 & 0.893 & 0.938 & 0.966  \\

Hyp ViT base fix $\tau$       & 0.826 & 0.898 & 0.937 & 0.965  \\
Hyp ViT base uncertainty as $\tau$  & 0.847 & 0.91 & 0.947 & 0.97  \\

Hyp ViT large fix $\tau$        & 0.838 & 0.905 & 0.945 & 0.969  \\
Hyp ViT large uncertainty as $\tau$ & 0.871 & 0.924 & 0.958 & 0.977 \\

\bottomrule
\end{tabular}
\end{table*}

\begin{table*}[ht]
\centering
\caption{The performance of Hyp ViT base uncertainty as $\tau$ on different scaling hyperparameters}\label{Tab:scale}
\begin{tabular}{l|llll}
\toprule
Datasets       & \multicolumn{4}{c}{CARS196} \\
\hline
Scale      & 1          & 2          & 4          & 8          \\
\hline
0.02       & 0.865 & 0.923 & 0.956 & 0.974 \\
0.04       & 0.865 & 0.922 & 0.953 & 0.974 \\
0.06       & 0.867 & 0.925 & 0.956 & 0.976  \\
0.08       & 0.869 & 0.925 & 0.955 & 0.974  \\
0.10       & 0.865 & 0.925 & 0.955 & 0.974  \\
0.12       & 0.869 & 0.924 & 0.953 & 0.973  \\
0.14       & 0.862 & 0.919 & 0.954 & 0.974  \\
0.16       & 0.858 & 0.918 & 0.951 & 0.971  \\
0.18       & 0.852 & 0.912 & 0.945 & 0.970  \\
\bottomrule
\end{tabular}
\end{table*}
\paragraph{configuration of the Training details}
The configuration of the training details of our algorithm on different datasets are presented in Table~\ref{Tab:configuration}.

The batch size configuration and number of samples per category in each batch are configured based on initial empirical results and other related methods.  

\paragraph{Comparison with other State-of-the-arts}
We first report and compare our experimental results with other State-of-the-arts on Table~\ref{Tab:sota}. On the table, Margin~\cite{8237571}, FastAP~\cite{cakir2019deep}, MIC~\cite{roth2019mic}, XBM~\cite{wang2020cross}, NSoftmax~\cite{zhai2018classification} are based on ResNet-50 backbone network. IRT~\cite{el2021training} is based on the DeiT backbone network. All the results are obtained from original papers. We include the more recent work that targets specifically Hyperbolic metric learning~\cite{ermolov2022hyperbolic}, and report their baseline results on ViT, DeiT, etc., backbone networks. The dimension of the feature embedding for all transformer-based models is 786-d. We transform the original feature embedding to 128-d hyperbolic embedding feature space and report the results.

We apply a similar baseline training loss function, i.e., Contrastive learning, to realize Hyperbolic metric learning. The key innovation of our method lies in applying the Hyperbolic embedding and its inherent uncertainty measurement in the metric learning process. We propose two adaptive metric learning approaches for Contrastive training and Triplet loss, respectively, fully utilizing the uncertainty measurement, thus realizing more effective and robust metric learning. 

As seen from Table~\ref{Tab:sota}, our methods incorporated with uncertainty measurement lead the baseline results reported in~\cite{ermolov2022hyperbolic} significantly. Especially, as the approaches implemented  in~\cite{ermolov2022hyperbolic} apply the Contrastive learning, instead of conventional margin-based metric learning, we would mainly compare ours based on the Contrastive learning with adaptive $tau$ with the results from~\cite{ermolov2022hyperbolic}. From Table~\ref{Tab:sota}, i.e., the second last line and line 18, our results lead the original Hyperbolic metric learning, by a large margin in the classical image retrieval task.

\paragraph{validation of the uncertainty measurement and uncertainty awareness} 
The uncertainty measurement is validated via the rising retrieval performance when applying our adaptive metric learning versus, the original metric learning scheme. The increase in final evaluation results actually validates two points: the uncertainty measurement is accurate and the proposed scheme of applying it in metric learning is effective. 

We perform a detailed ablation study reported in Table~\ref{Tab:Contrastive} and Table~\ref{Tab:triplet}, for adaptive Contrastive and Triplet loss, respectively. As can be observed from the tables, the performance of the original loss functions, when augmented with the uncertainty measurement and adaptive adjustment, is largely improved. This consistent phenomenon on different configurations validates the effectiveness of the uncertainty measurement. 

To validate the effectiveness of the uncertainty measurement and the uncertainty awareness in metric learning loss in a more explicit way, we additionally report the effect of uncertainty awareness of Contrastive loss, on different models, as shown in Table~\ref{Tab:testcase}. 

An ablation study on different $\tau$ values of the Contrastive loss on CUB200 datasets is illustrated in Table~\ref{Tab:abs_tau}. Though we could tune the values of $\tau$ to achieve comparable performance with the proposed Hyp-UML scheme, the consumed efforts and resource is much beyond the adaptive scheme.

\paragraph{Impact of the Curvature in Hyperbolic embedding.}
As stated in~\cite{9679192}, 
Hyperbolic curvature can measure the similarity between Hyperbolic geometry and Euclidean geometry. The work~\cite{9679192} also shows that different curvatures significantly affect distance metrics in Hyperbolic space. When the curvature decreases, the Hyperbolic embedded distance is more reflective of the tree structure because it is close to the shortest path length of the two nodes (i.e.,
Hyperbolic graph distance). When the curvature approaches zero, the Hyperbolic embedded distance is close to the Euclidean embedded distance, leading to the information loss
of the hierarchical structure.

However, constantly increasing the curvature only sometimes brings performance improvement. The curvature is more empirical, depending on the actual data structure. As a result, we performed an ablation study to check the impact of the curvature parameter on the final results.

In our experimental results, i.e., Table~\ref{Tab:Contrastive} and Table~\ref{Tab:triplet}, the curvature is not very sensitive Hyper-parameter, even though a suitable value still yields slightly better performance. For instance, a curvature value of 0.1 is relatively a more suitable choice in most of the cases, as can be observed from the tables. 
\begin{figure*}
    \centering
    \includegraphics[width=0.8\linewidth]{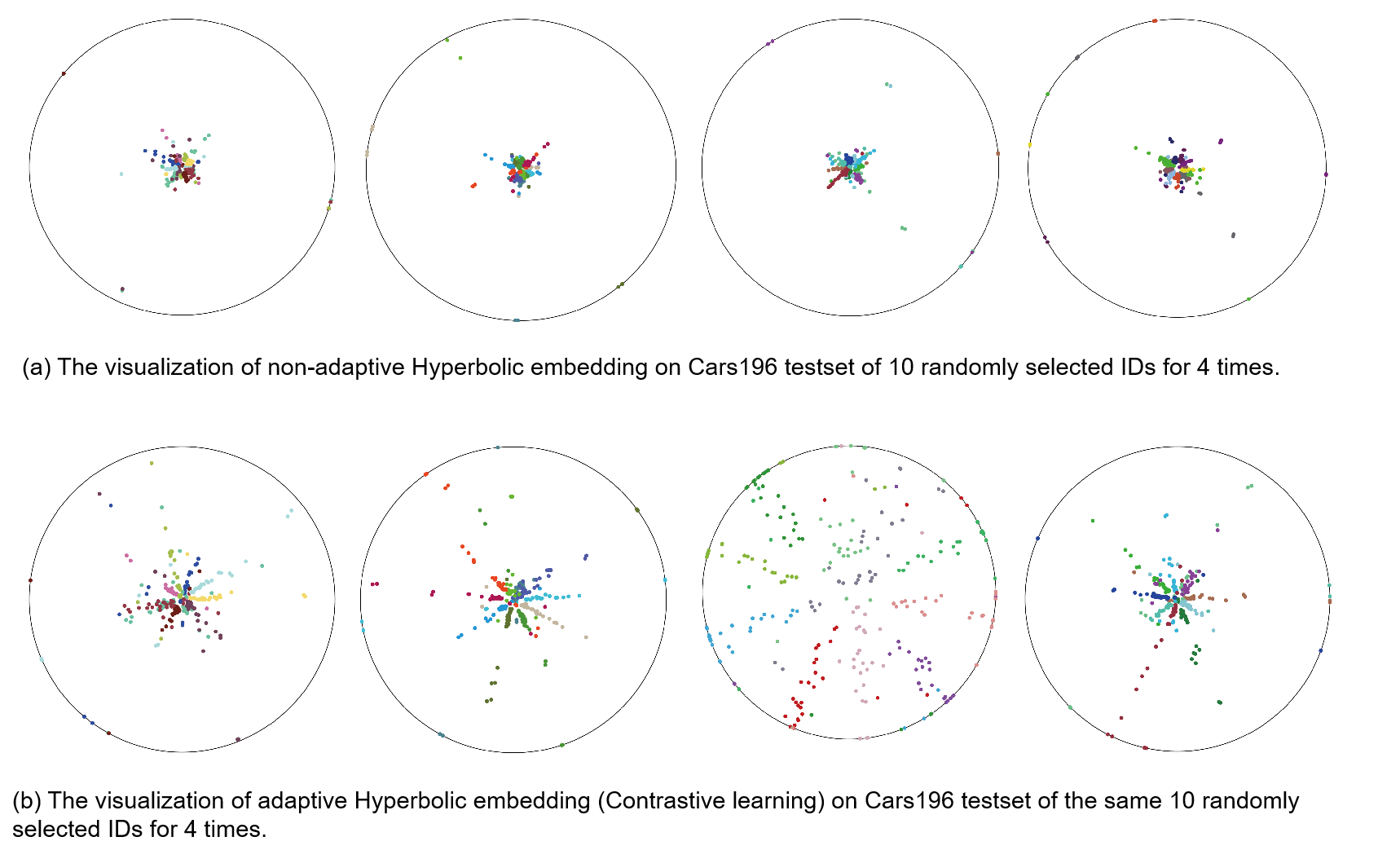}
    \caption{Visualization of the embeddings from Cars196 testset. (a) 10 randomly selected IDs for conventional embedding for 4 times. (b) The same 10 randomly selected IDs for Hyperbolic learning for 4 times, the model is trained via the proposed uncertainty-aware Contrastive learning.}
    \label{fig:vis}
\end{figure*}

\begin{figure}
    \centering
    \includegraphics[width=\linewidth]{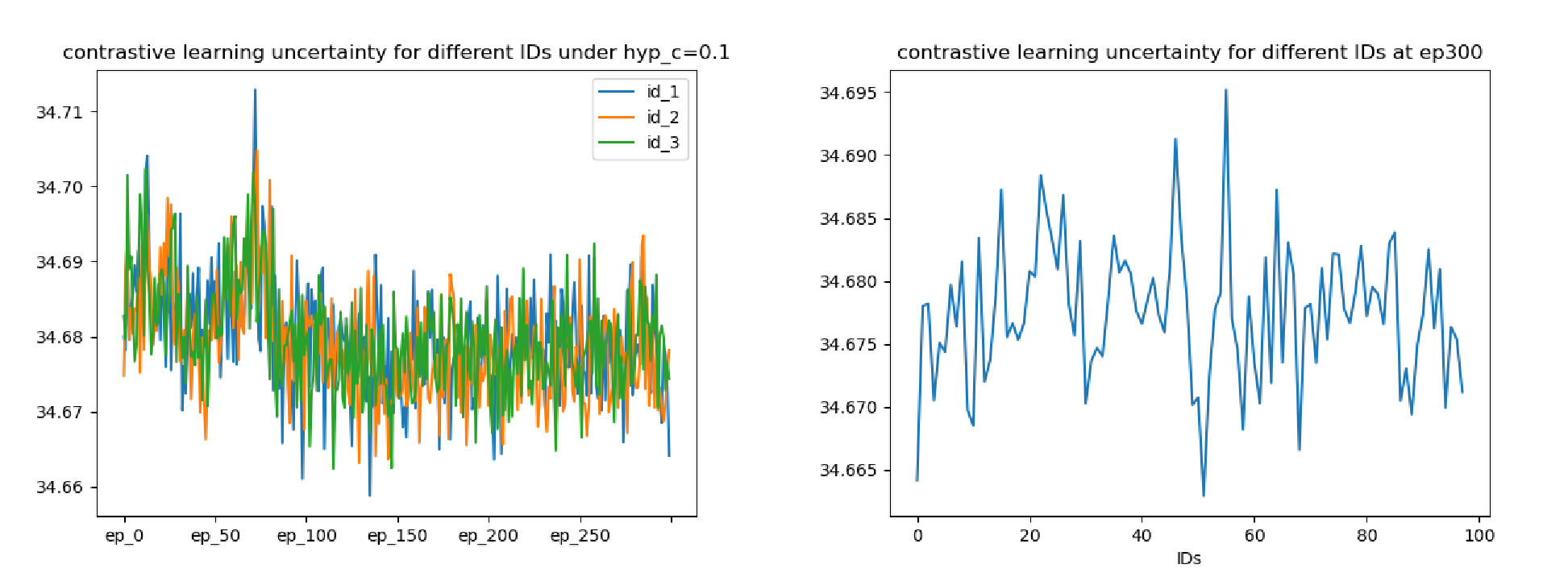}
    \caption{We visualize the values of uncertainties of the Contrastive learning scheme. The left figure plot the change of the uncertainties over training, it become more stable as the training is performed. The right figure show the uncertainty from different IDs at the same epoch. The uncertainty values are diverse, which shows that samples from different IDs have various uncertainty values. }
    \label{fig:vis_tau}
\end{figure}

\paragraph{The impact of the Hyperparameter $scale$ in Contrastive learning}
The impact of Hyperparameter $scale$ in Contrastive learning of ViT base model is reported in Table~\ref{Tab:scale}. The detailed results in the table on the different scaling factors, specifically, 
$\{0.02...0.18\}$ show that the scaling factor is not a sensitive Hyperparameter, though a suitable scaling factor can bring certain performance improvement. These experimental results validate the robustness of the proposed method.

\subsection{Qualitative Evaluations}
To evaluate the hyperbolic embedding qualitatively, we perform visualization of the Hyperbolic embedding with uncertainty awareness and the corresponding uncertainty 
 awareness embedding, as shown in Figure~\ref{fig:vis}. 

The top row of the figure is a visualization of uncertainty unawareness embedding from the BeiT model, trained via Contrastive loss. Due to the limitation of computing resources and more clarity in the visualization, we randomly selected 10 IDs from the Cars196 test set and selected them 4 times. The bottom row in the figure is a visualization of the proposed Hyperbolic uncertainty awareness embedding for the BeiT model, trained via the adaptive $\tau$ uncertainty-aware Contrastive loss scheme. 

The visualization of the embedding shown in the figure shows that the embedding with uncertainty awarded can better cluster the features in Poincaré's ball than the conventional method. It also shows that the embedding in Poincar\' e's ball can be well-distributed with a hierarchical structure, i.e., with different distances to the circle's boundary.

We visualize the values of uncertainty estimation of the Contrastive learning scheme in Figure~\ref{fig:vis_tau}. The left figure plot the change of the uncertainties over training, it become more stable as the training is performed. The right figure show the uncertainty from different IDs at the same epoch. The uncertainty values are diverse, which shows that samples from different IDs have various uncertainty values.

The visualization is performed via this procedure: we first extract the feature embedding from the model on the test set. Subsequently, we apply the HoroPCA~\cite{chami2021horopca} to reduce the dimension of the embedding (128-d) to 32-d. Subsequently, we apply the CO-SNE~\cite{guo2022co} to reduce the dimension to 2-d and then visualize the final embedding in Poincar\' e's ball. HoroPCA is a general method for dimension reduction for Hyperbolic embedding, and CO-SNE is a recently developed visualization tool for Hyperbolic space.

\section{Conclusion}
Conventional deep Neural Networks and their corresponding popular metric learning algorithms, e.g., Constrasive training and various margin-based losses, are all in Euclidean distance space. Recent research shows the limiting aspects of Euclidean embedding, especially in hierarchical structural data representation learning. On the contrary, Hyperbolic embedding, a Riemannian embedding method, has an inherent advantage in representing Hierarchical data structure, due to its exponential transformation. In this paper, we propose a Hyperbolic embedding method for Transformer models to solve the critical difficulties of image retrieval. In short, we transform the features from a conventional Transformer model to Hyperbolic space, bypassing the complex algorithm of Hyperbolic Neural Networks. Second, we propose an uncertainty measurement scheme for hyperbolic image embedding and incorporate it into two types of metric learning losses, i.e., Constastive learning and margin-based metric learning. The advancement in numerical results and certain qualitative evaluation validate the effectiveness of the proposed method. A comprehensive ablation study evaluates each component of the method, validating the superiority of the proposed algorithm in image retrieval. Future work includes dynamic curvature learning for Hyperbolic embedding. 

\bibliographystyle{ieeetr}
\bibliography{shiyang}

 





\end{document}